\documentclass[10pt,conference,a4paper]{IEEEtran}

\IEEEoverridecommandlockouts
\usepackage[utf8]{inputenc}
\usepackage{cite}    
\usepackage{amsmath,amssymb,amsfonts}
\usepackage{graphicx}
\usepackage{textcomp}
\usepackage{xcolor}
\usepackage{balance}
\usepackage{algorithm,algcompatible,amsmath}    
\usepackage[noend]{algpseudocode}               
\algnewcommand\INPUT{\item[\textbf{Input:}]}
\algnewcommand\OUTPUT{\item[\textbf{Output:}]}
\def\BibTeX{{\rm B\kern-.05em{\sc i\kern-.025em b}\kern-.08em
    T\kern-.1667em\lower.7ex\hbox{E}\kern-.125emX}}
\begin{document}
\title{Why Layer-Wise Learning is Hard to Scale-up and a Possible Solution via Accelerated Downsampling
}

\author{\IEEEauthorblockN{Wenchi Ma$^\dag$, Miao Yu$^\dag$, Kaidong Li$^\dag$, Guanghui Wang$^{\dag\ddag}$}
\IEEEauthorblockA{$^\dag$ \textit{Electrical Engineering and Computer Science},
\textit{The University of Kansas,
Lawrence, KS, USA 66045}\\
$^\ddag$ \textit{Department of Computer Science},
\textit{Ryerson University,
Toronto, ON, Canada M5B 2K3}\\
\{wenchima, ghwang\}@ku.edu
}}

\maketitle

\begin{abstract}
Layer-wise learning, as an alternative to global back-propagation, is easy to interpret, analyze, and it is memory efficient. Recent studies demonstrate that layer-wise learning can achieve state-of-the-art performance in image classification on various datasets. However, previous studies of layer-wise learning are limited to networks with simple hierarchical structures, and the performance decreases severely for deeper networks like ResNet. This paper, for the first time, reveals the fundamental reason that impedes the scale-up of layer-wise learning is due to the relatively poor separability of the feature space in shallow layers. This argument is empirically verified by controlling the intensity of the convolution operation in local layers. We discover that the poorly-separable features from shallow layers are mismatched with the strong supervision constraint throughout the entire network, making the layer-wise learning sensitive to network depth. The paper further proposes a downsampling acceleration approach to weaken the poor learning of shallow layers so as to transfer the learning emphasis to deep feature space where the separability matches better with the supervision restraint. Extensive experiments have been conducted to verify the new finding and demonstrate the advantages of the proposed downsampling acceleration in improving the performance of layer-wise learning.
\end{abstract}


\section{Introduction}
Deep Convolutional Neural Networks (CNNs) trained on large-scale supervised data, with the prediction error backpropagating layer by layer from the output layer to bottom hidden layers, has become the mainstream approach in most computer vision tasks \cite{b21} \cite{b20}. This has motivated successes in the applications of image classification \cite{b22} \cite{b23}, object detection and segmentation \cite{b24} \cite{b25} \cite{b1} \cite{b2} \cite{b26} \cite{b39} \cite{b40}, speech recognition \cite{b3}, meta-learning \cite{b27}, and reinforcement learning \cite{b4}, etc. Up to now, there is no solid schema to excellently and wide-applicably define architecture or training procedures for certain tasks, although we have architecture searching as assistance \cite{b5}, which is tedious and inefficient. The essence here is we can not handle each layer from design to operation. It is not clear how the individual intermediate layer contributes to the final functional behavior of the network and how these layers work together for the prediction performance. 

\begin{figure}
    \includegraphics[width=1.0\linewidth]{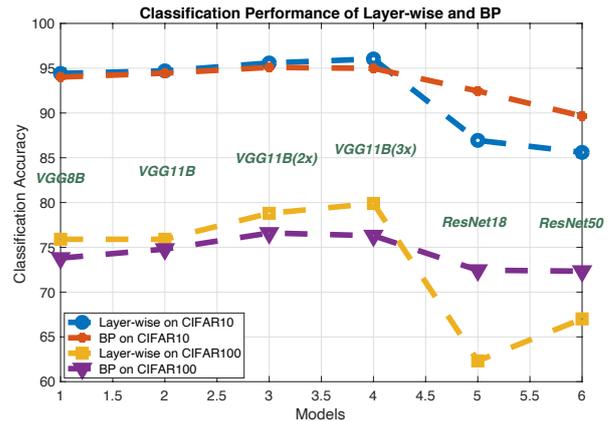}\hfill
    \vspace{-2mm}
    \caption{Classification performance between layer-wise and global learning.}
	\label{fig:compare}
 	\vspace{-5mm}
\end{figure}

There are other potential issues worthy to be considered. The weights of hidden layers can only be updated after a whole forward and backward pass. This backward locking prevents the parallelization of parameter updates. It is also inefficient in terms of memory resources to store the layer activation of the entire network, failing for memory reuse. Several works have proposed to avoid the backward locking and memory reuse problem, such as modeling synthetic gradients replacing back propagated error gradients or exactly reconstructing the activation of current layers from the next layer's so as to avoid storing the activation of most layers in memory during back propagation \cite{b28} \cite{b29}. While most of these algorithms are limited at the theoretical level, simple network architectures, or small datasets. As is shown in Figure \ref{fig:compare}, where we list and compare layer-wise learning with corresponding global BP algorithms on CIFAR10 and CIFAR100 respectively. Here the layer-wise learning models employ Predsim loss \cite{b6} for supervised constraint which has obtained state-of-the-art performance on general benchmarks. It appears that the state-of-the-art layer-wise approach works well for shallow networks, such as the VGG family, while when it comes to ResNet models with a deeper scale, layer-wise learning is hard to be competitive with global BP. 

Recent researches demonstrate that the backward locking can be effectively avoided by local-error learning \cite{b6} \cite{b8}\cite{b11} \cite{b10} \cite{b7}. It enables memory reuse and local performance supervision by training with locally generated errors. The local loss relies much less on accessing a full gradient. The gradient is not back propagated to previous layers, but the weights of hidden layers are updated in the forward pass. This enables greedily training layers one at a time. In this way, we can specify the objective for each layer by the classic learning principle, which encourages refining specific representation properties \cite{b15}, such as progressive linear separability \cite{b30} \cite{b8}. 

At the inference stage, the network behaves in a standard global procedure. Unfortunately, recent works have not convincingly demonstrated that local-error training can be scaled to deeper form and tackle large-scale problems \cite{b11} \cite{b31} \cite{b32} \cite{b33} \cite{b34} \cite{b10} \cite{b7}. Greedy layerwise learning scales to ImageNet, realizing the same accuracy to the VGG-11 by introducing the by-pass CNN branch before linear classification after each convolutional layer \cite{b8}. Nøkland and Eidnes \cite{b6} extends the single-layer sub-networks by proposing two different supervised loss functions to generate local error signals for the hidden layers, and obtains competitive or even better results on most popular image classification datasets. However, all these works are confined to simple network architectures, at most to VGG19. There is no study to explore and reveal why current state-of-the-art layer-wise learning strategies can not be scaled to larger networks, such as the ResNet family.  

In this context, we demonstrate that the fundamental reason that impedes the layer-wise learning towards a deeper and larger scale lies in the poor separability of the bottom layers. By quantitatively showing and analyzing the performance of each hidden layer during the training process, we have found that the network as a whole is very sensitive to the separability of bottom layers, the initial classification ability of the network. Since layer-wise learning equips each layer with independent training, the progressive linear separability encounters a big challenge due to the unmatched supervision constraints. Based on the above observation and analysis, our contributions in this paper can be concluded as below. 
\begin{itemize}
    \item We analyze the separability of feature space along with the training procedure from the perspectives of both time and spatial domains. By extensive controlling experiments, we find that the fundamental reason behind the difficulty of scaling layer-wise learning towards deeper and larger networks is the poor separability of bottom layers, especially in the situation that the consistent supervision constraint is devoid. 
    \item We propose a downsampling acceleration scheme to boost the interpretation for feature space and match it with the supervision constraints for earlier layers so as to enhance the classification performance of the network. The scheme is proved to be highly-effective and comparative to global back propagation.
    \item We demonstrate that efficiently utilizing the interpretation ability of feature space is an effective method for the current state-of-the-art layer-wise regime. The study provides a perspective of solving the current difficulty of layer-wise learning and inspires further study on this problem. 
\end{itemize}

\section{Related Work}
Layer-wise learning, in early stage, has been explored to work as the initialization of networks for subsequent end-to-end supervised learning, such as the work of Bengio et al. \cite{b41}, where the local loss functions were used to pre-train the hidden layers independently of the global loss. While this has only been proved in certain cases that can improve the performance after fine-tuning with global backpropagation \cite{b8}. Then, there are works on large-scale supervised learning show that the modern existing training techniques enable replacing the layerwise initialization \cite{b9}. 

Some other works were motivated by the difficulties of gradients vanishing and explosion. Shalev-Shwartz and Malach trained the network using the layer-wise learning fashion and showed that their work can be provably generalized to the image models with a restricted class \cite{b10}. Huang et al. \cite{b11} introduced the boosting theory in the residual network \cite{b12} to sequentially train layers. Marquez et al. \cite{b13} and Kulkarni \& Karande \cite{b14} set strict and limited experimental conditions for layer-wise training. However, experimental results demonstrated that these layer-wise learning approaches are limited for datasets and hard to obtain competitive results with the end-to-end global back propagation. Eugene et al. \cite{b8} proposed the greedy layer-wise learning and proved it can be scaled to ImageNet, which relies on a simpler objective function and can be competitive with end-to-end approaches. In their work, good results were first reported on CIFAR10 with the error of 7.2\% using the local classifiers and ensembles, approaching the accuracy of global back propagation. Nøkland and Eidnes \cite{b6} contributed to this community by showing that the local classifiers combined with the local similarity matching loss can match the global back-propagation learning in terms of the test error. Their work obtains the state-of-the-art results on VGG networks and outperforms the corresponding BP algorithm in the same experimental settings. However, all these works realize the state-of-the-art results in limited conditions. In terms of network scales, none of the existing works solve the problem of scaling the current layer-wise learning to deeper structure, such as the ResNet family.  

Other related research threads include adding layers to the base networks and then train the networks in the end-to-end fashion. Brock et al. proposed to build a network in stages, in which some layers are progressively frozen so as to speed up the training procedure \cite{b17}. Mosca and Magoulas \cite{b18} proposed to progressively stack layers, training the end-to-end network step by step. In our work, we do not discuss performing fine-tuning for the entire network. Another thread is using the local loss functions to train sub-networks to approximate the true gradient, which is a method to evade the backward locking problem \cite{b19}. Specifically, the true gradient is predicted from the layer above by training the synthetic gradient models using the L2 loss function. This work is related to the layer-wise learning as the hidden layer is trained by the labeled target information. Different from previous work, we discuss the layer-wise problem not from approximating the true gradient, we employ the labeling information to build the error regression independent of adjacent layers. 

\section{Essential Reason for Development Dilemma}
In this section, we formalize the analysis towards the essential reason why the current state-of-the-art layer-wise regime is unable to adapt to larger and deeper scales and show quantitative results of the separability layer-by-layer in both spatial and time domains. Then, we verify the above analysis using popular models that have top performances with layer-wise learning and draw the conclusion that the poor separability of bottom layers is the essential reason that limits the scalability of the layer-wise learning.

\subsection{Architecture Formulation and Progress}
The state-of-the-art hierarchical architecture of local-error learning adopts the standard convolution operation and fully connection \cite{b8} \cite{b6}. Instead of global back propagating loss error, these models are trained in succession, where each weight layer is trained by the corresponding local error signal, back propagated down this single unit. In this paper, we focus on the layer-wise CNN regime as they are the mainstream local-error learning scheme that has caused extensive attention and proved to be robust on popular benchmarks. The architecture has $N$ operation units. Each unit is composed of a convolution layer in the backbone network (base convolution unit) and an auxiliary or by-pass classification branch. The branch serves as a local classifier that provides the local learning signal, in the form of sub-networks. Each feature output $O_{j}$ of the convolution unit $j$ in the base network will be fed into the auxiliary classification branch to obtain local prediction $c_{j}$. Then, $O_{j}$ will be fed to the next convolution unit $j+1$ in base network, as the input $x_{j+1}$. 
\begin{equation}\label{eq1}
x_{j+1}=O_{j}=W_{\theta_{j}}x_{j}
\end{equation}
\begin{equation}\label{eq2}
c_{j}=C_{\gamma_{j}}P_{j}W_{\theta_{j}^{'}}x_{j+1}
\end{equation}
where $c_{j}$ is the classification output of layer $j$. $W_{\theta_{j}}$ is the weight parameters of convolution operation unit $j$ in the base network with all the parameters denoted by $\theta_{j}$. $W_{\theta_{j}^{'}}$ is the parameters of the convolution layer in the auxiliary classification branch, corresponding to layer $j$. $P_{j}$ is the pooling operator for down-sampling. $C_{\gamma_{j}}$ is the local classifier with its parameters $\gamma_{j}$. The combination of $C_{\gamma_{j}}P_{j}W_{\theta_{j}^{'}}$ is the popular structure of the local classifier branch in recent layer-wise works, which has shown robust performances in various situations \cite{b8} \cite{b6}. 
Compared with the global BP in equation \ref{eq1}, layer-wise learning updates the layer weights by the prediction result of the current local classifier so that the weight parameters can complete update before the feature outputs being fed to the next convolution unit of the base network. 
\begin{equation}
c=C_{\gamma}P_{k}W_{\theta_{k}}...P_{2}W_{\theta_{2}}P_{1}W_{\theta_{1}}P_{0}W_{\theta_{0}}x_{0}
\end{equation}
where we suppose that there are $k$ convolution operation units in the entire base network and each unit consistently consists of convolution $W_{\theta}$ and spatial pooling $P$.

Based on the above analysis, the training procedure for the local-error learning could be layer-wise. Specifically, the training for layer $j$ is independent, keeping all other parameters of the remaining layers fixed. $\theta_{j}$ is updated by optimizing ${\theta_{j}, \theta_{j}^{'}, \gamma_{j}}$, so as to obtain the best classification performance of the auxiliary classifier $C_{\gamma_{j}}$ \cite{b8}. For a training set of \{$\mathbf{X}, \mathbf{y}$\}, we try to minimize the empirical risk of classification \cite{b8}: 
\begin{equation}
\hat{R}(f; \theta, \theta^{'}, \gamma) = \sum l(f(\mathbf{X};\theta, \theta^{'}, \gamma), \mathbf{y})
\end{equation}
where the mapping $f(\mathbf{X};\theta, \theta^{'}, \gamma)$ is parameterized by \{$\theta, \theta^{'}, \gamma$\} and optimized by the loss $l$. Thus, for a certain depth $j$, the layer-wise learning algorithm takes $f_{j} = f(\mathbf{x}_{j};\theta_{j}, \theta_{j}^{'}, \gamma_{j})$ so that the local optimization procedure aims to minimize the rick of $\hat{R}(f_{j}; \theta_{j}, \theta_{j}^{'}, \gamma_{j})$. This local learning procedure is to train the classifier $C_{\gamma j}$ on top of its input $\mathbf{x}_{j}$ to obtain the optimized $\hat{\theta}_{j}$, as shown in the following algorithm. 
\begin{algorithm}
\caption{Layer-Wise Learning}
\begin{algorithmic}[1]
    \INPUT Training data sample \{$\mathbf{X}, \mathbf{y}$\}
    \State $j=1$
    \While{j $\leq$ N}  
        \State $x_{j+1}=O_{j}=W_{\theta_{j}}x_{j}$
        \State $(\hat{\theta}_{j}, \hat{\theta}^{'}_{j}, \hat{\gamma}_{j}) = \arg\min_{\theta_{j}, \theta^{'}_{j}, \gamma_{j}} \hat{R}(f_{j};\theta_{j}, \theta^{'}_{j}, \gamma_{j})$
        \State $j+=1$
    \EndWhile  
\end{algorithmic}
\end{algorithm}

The performance of the cascaded training progress is hinged on the individual property of each local problem, the optimization of each layer. Thus, the sufficient condition of Algorithm 1 that would converge to the optimal solution is the local problem approaching the optimal.

\subsection{Separability of Feature Space}
Based on the above analysis, the classification performance of each certain layer unit contributes to the ultimate property of the entire network, where the feature space separability could be one essential determinant. This means the progress of feature space reflected in the variation of separability would determine the optimization results of the layer-wise learning. From this perspective, the consistency of feature space separability throughout each layer unit of the entire network would vary the final performance of layer-wise networks. In the following, we explore the essential reason that prevents the layer-wise learning from generalizing to deeper scales. 

From Figure \ref{fig:compare}, we can see that the classification performance of layer-wise learning is impaired with the increase of the network depth, and the gap between layer-wise and global learning is increased as well at the same time. As shown in Figure \ref{fig:compare}, the current state-of-the-art layer-wise learning models are based on VGG frameworks. While for those based on ResNet frameworks, they usually fall behind their counterparts with global BP, even though their network depth is similar to that with VGG-based frameworks. Thus, we first analyze the hierarchical structures of VGG and ResNet and then explore the reason that gives rise to the performance difference in terms of layer-wise learning. 

\begin{figure}
    \includegraphics[width=1.0\linewidth]{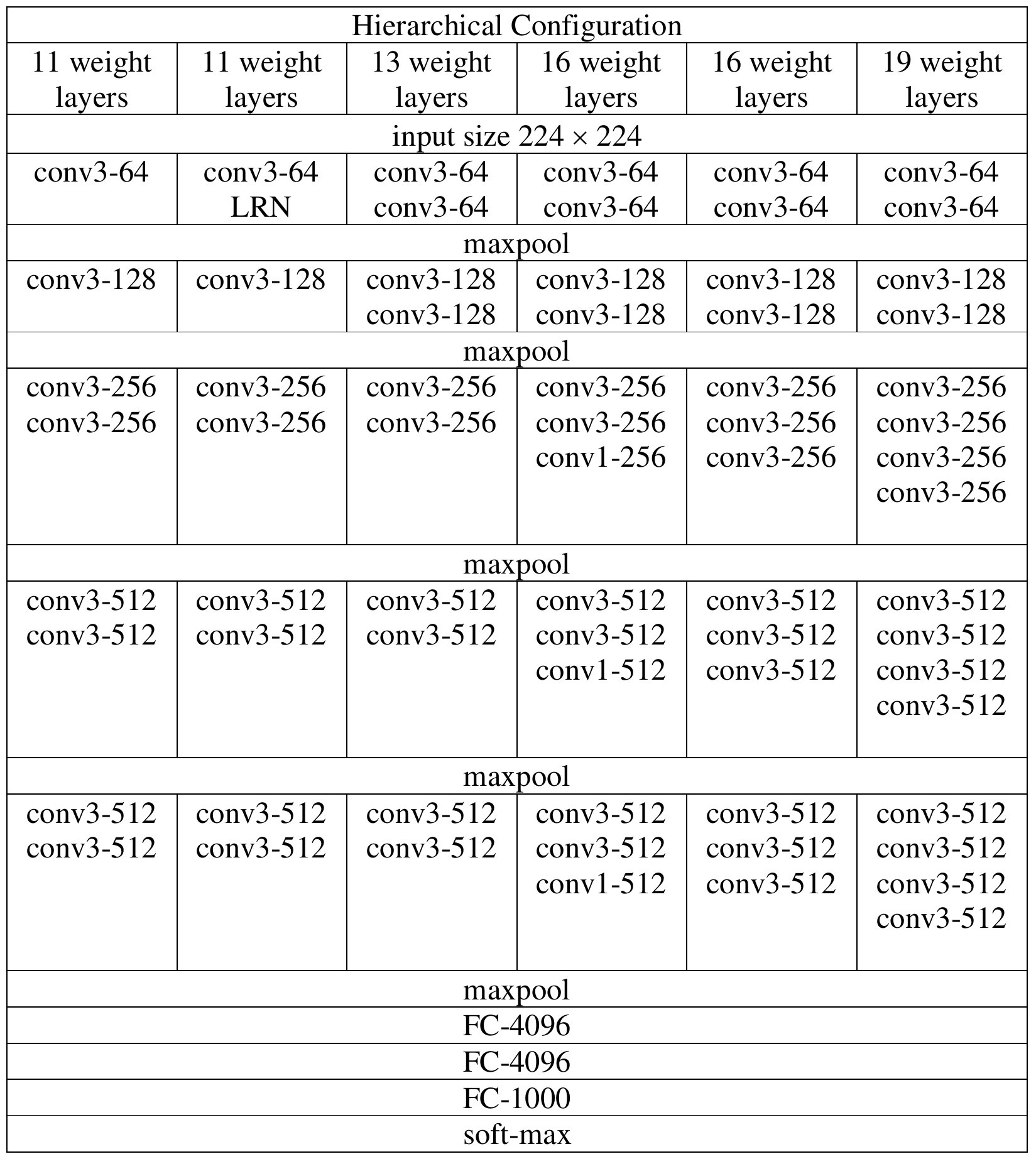}\hfill
    \vspace{-2mm}
    \caption{Hierarchical structures of VGG family. }
	\label{fig:vgg}
 	\vspace{-3mm}
\end{figure}

\begin{figure}
    \includegraphics[width=1.0\linewidth]{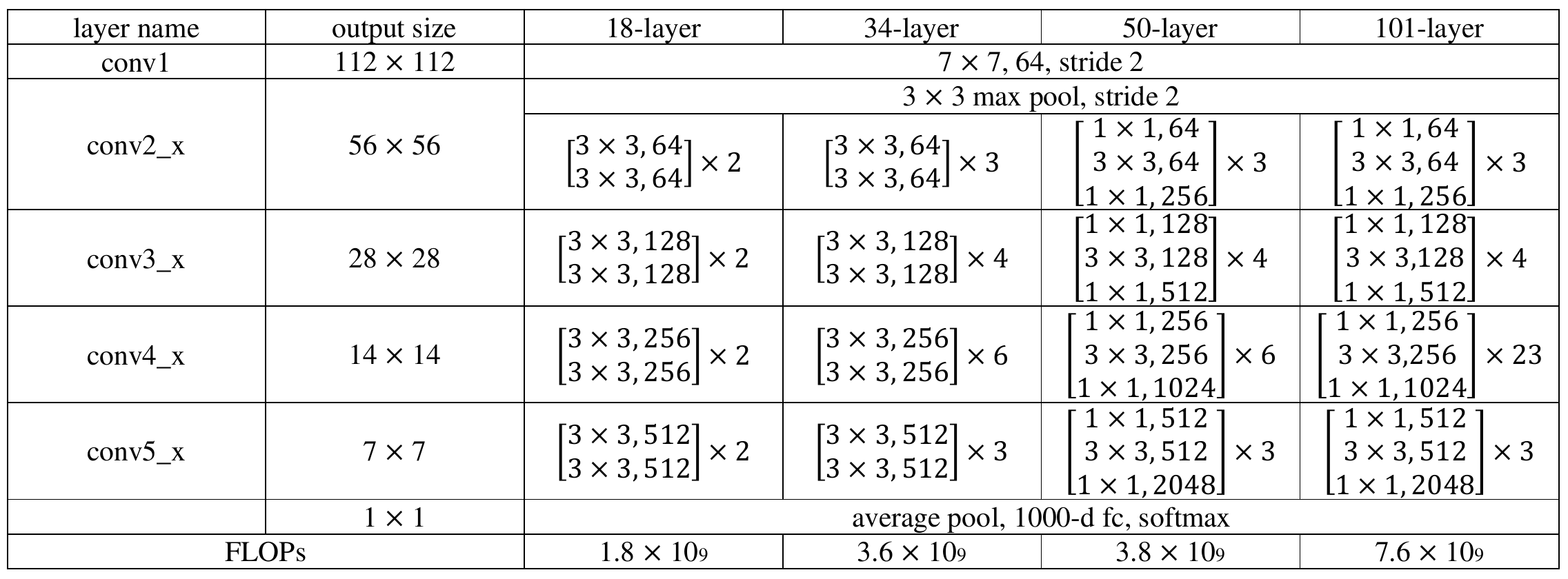}\hfill
    \vspace{-2mm}
    \caption{Hierarchical structures of ResNet family. }
	\label{fig:resnet}
 	\vspace{-3mm}
\end{figure}

As shown in Figure \ref{fig:vgg}, all VGG models perform max pooling (/2) after every two consecutive convolutions within the first two down-sampling operations. This means VGG models reduce the feature map resolution to its 1/4 after only four convolutions. For the remaining layers, VGG11 and VGG13 remain the same down-sapling frequency of every other two convolutions, while VGG16 and VGG19 increase the frequency to 3 and 4 respectively. In contrast, ResNet models keep the down-sampling frequency of every 4 convolutions or more after the first max-pooling as shown in Figure \ref{fig:resnet}. On average, the number of convolution operations at the feature map sizes, 56, 28, and 14 for ResNet models is two to three times of that for VGG models

Based on the fact that the interpretability of feature space increases with the resolution of feature maps getting diminished and feature expression getting highly abstract \cite{b2} \cite{b38}, the strong supervised classification constraint towards the feature space with relatively high resolutions is unmatched. To which, the layer-wise learning could be more sensitive as a lack of consistent constraint compared with global learning algorithms. In view of this, we believe that relatively intensive convolution operation on poor interpretable features would lead to optimization deviation, especially for the situation when missing effective consistency constraint. Thus, the parameter update suffers during the learning procedure of local layers, which could definitely impair the ultimate training performance of the network. We argue that this is the fundamental reason why the layer-wise learning is sensitive to the network hierarchy and depth. In the following part, we verify this argument by controlling the intensity of convolution operation, especially in shallow layers where the feature space is usually poorly interpretable and separable \cite{b38}.

\subsection{Essential Reason}
We explore the inner relation between the classification performance of layer-wise learning and the separability of feature space by setting the intensity of convolution operation. We introduce the VGG family and ResNet family as the base networks and verify the above relationship on popular classification benchmarks, CIFAR10 and CIFAR100. We retain the specific model with the same learning hyper-parameters, including solver, learning rate, drop out, etc. We only set different intensities of convolution for different layer units. For the ``Hierarchy" term in the following tables, each number represents the number of convolution layers between pooling layers. Thus, the number list for each model shows the hierarchical structure of the network. We generally take the first half rows of numbers in the list as the indication of shallow layers' hierarchy and the latter half rows refer to the hierarchy of deep layers. For each specific setting (base model, learning parameters, hierarchy), the classification performances of both layer-wise and global learning are provided. 

\setlength{\tabcolsep}{4pt}
\begin{table}
\begin{center}
\caption{Classification Verification with VGGs on CIFAR10.}
\label{table:vgg_cifar10}
\begin{tabular}{|c|c|c|c|c|c|}
\hline
Model & LR & DR & Hierarchy & Layer-wise & Global\\
\hline
VGG8 & 5e-4 & 0.2 &[1,1,1,1,1] & 89.72 & 90.59\\
\hline
VGG11 & 5e-4 & 0.2 & [1,1,2,2,2] & 90.6 & 91.6\\
\hline
VGG16 & 3e-4 & 0.25 & [2,2,3,3,3] & 89.2 & 92.03\\
\hline
VGG16\_e1 & 3e-4 & 0.25 & [3,3,4,4,4] & 87.69 & 92.5\\
\hline
VGG16\_e2 & 3e-4 & 0.25 & [4,4,5,5,5] & 86.2 & 91.72\\
\hline
VGG16\_e3 & 3e-4 & 0.25 & [5,5,6,6,6] & 84.3 & 90.67\\
\hline
VGG19 & 3e-4 & 0.25 & [2,2,4,4,4] & 89.14 & 92.21\\
\hline
VGG19\_e1 & 3e-4 & 0.25 & [3,3,5,5,5] & 87.77 & 91.93\\
\hline
VGG19\_e2 & 3e-4 & 0.25 & [4,4,6,6,6] & 86.0 & 91.65\\
\hline
VGG19\_e3 & 3e-4 & 0.25 & [5,5,7,7,7] & 83.7 & 89.83\\
\hline
VGG8b & 5e-4 & 0.2 & [2,2,1,1] & 94.59 & 94.15\\
\hline
VGG8b\_e1 & 5e-4 & 0.2 & [3,3,2,2] & 95.05 & 95.0\\
\hline
VGG8b\_e2 & 5e-4 & 0.2 & [4,4,3,3] & 94.7 & 95.2\\
\hline
VGG8b\_e3 & 5e-4 & 0.2 & [5,5,4,4] & 94.52 & 95.1\\
\hline
\end{tabular}
\end{center}
\end{table}
\setlength{\tabcolsep}{1.4pt}

\setlength{\tabcolsep}{4pt}
\begin{table}
\begin{center}
\caption{Classification Verification with VGGs on CIFAR100.}
\label{table:vgg_cifar100}
\begin{tabular}{|c|c|c|c|c|c|}
\hline
Model & LR & DR & Hierarchy & Layer-wise & Global\\
\hline
VGG8 & 5e-4 & 0.2 &[1,1,1,1,1] & 65.38 & 65.82\\
\hline
VGG11 & 5e-4 & 0.2 & [1,1,2,2,2] & 67.88 & 65.84\\
\hline
VGG16 & 3e-4 & 0.1 & [2,2,3,3,3] & 66.15 & 68.25\\
\hline
VGG16\_e1 & 3e-4 & 0.1 & [3,3,4,4,4] & 64.43 & 70.13\\
\hline
VGG16\_e2 & 3e-4 & 0.1 & [4,4,5,5,5] & 61.35 & 69.52\\
\hline
VGG16\_e3 & 3e-4 & 0.1 & [5,5,6,6,6] & 57.98 & 67.93\\
\hline
VGG19 & 3e-4 & 0.15 & [2,2,4,4,4] & 65.76 & 69.4\\
\hline
VGG19\_e1 & 3e-4 & 0.15 & [3,3,5,5,5] & 63.88 & 70.55\\
\hline
VGG19\_e2 & 3e-4 & 0.15 & [4,4,6,6,6] & 60.58 & 68.95\\
\hline
VGG19\_e3 & 3e-4 & 0.15 & [5,5,7,7,7] & 57.22 & 65.68\\
\hline
VGG8b & 5e-4 & 0.05 & [2,2,1,1] & 73.93 & 73.47\\
\hline
VGG8b\_e1 & 5e-4 & 0.05 & [3,3,2,2] & 74.5 & 74.05\\
\hline
VGG8b\_e2 & 5e-4 & 0.05 & [4,4,3,3] & 74.2 & 71.79\\
\hline
VGG8b\_e3 & 5e-4 & 0.05 & [5,5,4,4] & 73.02 & 69.0\\
\hline
\end{tabular}
\end{center}
\end{table}
\setlength{\tabcolsep}{1.4pt}

If we observe the layer hierarchy of VGG8, VGG11, VGG16, and VGG19 in Table \ref{table:vgg_cifar10} and Table \ref{table:vgg_cifar100}, it is clear that their convolution intensity is increasing with the network going deeper. VGG11 keeps the same number of convolution layers within the first two pooling layers while it increases the convolution operations by 1 between every two pooling layers in the later part. Thus, VGG11 theoretically has a better interpretation towards deep features that are highly-abstract and semantic. The deeper local network brings in stronger semantic expression and matches the better separability of deep features produced in deep layers. 

From the results on CIFAR10 and CIFAR100, VGG11 leads VGG8 by around 1\% in both the layer-wise and global learning cases. This is similar to VGG16 and VGG19, which have the same convolution hierarchy in the shallow part while VGG19 increases the convolutional operations in the deep part. Their results follow the same pattern with that of VGG8 and VGG11. While if we compare the models with the same deep layer hierarchy but with different shallow layer distribution, such as VGG16\_e1 and VGG19, VGG19\_e1 and VGG16\_e2 in Table \ref{table:vgg_cifar10} and Table \ref{table:vgg_cifar100}, it can be found that models with fewer shallow convolutions tend to yield higher classification accuracy. VGG16\_e1 surpasses VGG19 by 1-2\% on average and VGG19\_e1 leads VGG16\_e2 by the same margin on both CIFAR10 and CIFAR100. 

This phenomenon applies to the ResNet family as well. From Table \ref{table:resnet_cifar10} and Table \ref{table:resnet_cifar100}, ResNet50, ResNet50\_e1, and ResNet50\_e2 share the same deep layer hierarchy while their convolution operations in shallow layers are in an increasing number. Their classification performances, on the opposite, are degenerated. Thus, although ResNet18 and its variations increase convolution operations between every two pooling layers, getting deeper, it seems the classification performance is more sensitive to the shallow hierarchy. These observations reveal the fact that the interpretive ability of features produced in shallow layers has a big impact on the final performance of the layer-wise learning. Massive convolution operations on features with poorly-semantic expression leads to mismatched learning procedure, resulting in deviated parameters update and learning regression. 

Based on the above theoretical and experimental analysis, we can draw the conclusion that the poor separability of features produced in shallow layers is the main reason that impedes the layer-wise learning getting a deeper scale. 

\setlength{\tabcolsep}{4pt}
\begin{table}
\begin{center}
\caption{Classification verification with ResNets on CIFAR10.}
\label{table:resnet_cifar10}
\begin{tabular}{|c|c|c|c|c|c|}
\hline
Model & LR & DR & Hierarchy & Layer-wise & Global\\
\hline
ResNet18 & 3e-4 & 0.3 &[2,2,2,2] & 86.94 & 92.45\\
\hline
ResNet18\_e1 & 3e-4 & 0.3 & [3,3,3,3] & 83.64 & 91.62\\
\hline
ResNet18\_e2 & 3e-4 & 0.3 & [4,4,4,4] & 79.84 & 89.98\\
\hline
ResNet50 & 3e-4 & 0.3 & [3,4,6,3] & 85.6 & 89.65\\
\hline
ResNet50\_e1 & 3e-4 & 0.3 & [4,5,6,3] & 83.64 & 88.52\\
\hline
ResNet50\_e2 & 3e-4 & 0.3 & [5,6,6,3] & 81.23 & 87.83\\
\hline
\end{tabular}
\end{center}
\end{table}
\setlength{\tabcolsep}{1.4pt}

\setlength{\tabcolsep}{4pt}
\begin{table}
\begin{center}
\caption{Classification verification with ResNets on CIFAR100.}
\label{table:resnet_cifar100}
\begin{tabular}{|c|c|c|c|c|c|}
\hline
Model & LR & DR & Hierarchy & Layer-wise & Global\\
\hline
ResNet18 & 3e-4 & 0.1 &[2,2,2,2] & 62.32 & 72.45\\
\hline
ResN18\_e1 & 3e-4 & 0.1 & [3,3,3,3] & 62.85 & 71.38\\
\hline
ResNet18\_e2 & 3e-4 & 0.1 & [4,4,4,4] & 59.98 & 69.74\\
\hline
ResNet50 & 3e-4 & 0.1 & [3,4,6,3] & 67.01 & 72.34\\
\hline
ResNet50\_e1 & 3e-4 & 0.1 & [4,5,6,3] & 66.06 & 70.73\\
\hline
ResNet50\_e2 & 3e-4 & 0.1 & [5,6,6,3] & 65.42 & 70.97\\
\hline
\end{tabular}
\end{center}
\end{table}
\setlength{\tabcolsep}{1.4pt}

\section{Separability Enhancement and Accelerated Downsampling}
In order to improve the performance of layer-wise learning at deeper scales, we propose corresponding scheme to overcome its limitations. 

\subsection{Separability Enhancement}
Current layer-wise learning networks literally follow the pattern of global back propagation, learning from bottom layers to top layers. However, layer-wise learning heavily depends on the learning procedure of each local layer. It is well-known that the feature space in shallow layers is usually with sufficient feature details while lacking of effective semantic expression and separability. In global learning, to deepen the network layer distribution is very effective as the feature outputs of deeper layers could become highly-abstract and semantic, which makes the learning process much easier with the consistent constraint provided by the global back propagation. Given a list of convolution layer activations $H=[\mathbf{h_{1}, h_{2},\dots, h_{n}}]$, where $n$ refers to the layer depth of the entire network, and the label matrix $Y=(\mathbf{y_{1}, y_{2}, \dots, y_{n}})$. We have the global learning loss, 
\begin{equation}
\begin{aligned}
L_{G} = ||f(\mathbf{h_{n}})-Y||_{F}^{2}
\end{aligned}
\end{equation}
where $f$ is the classification mapping of the network.
In layer-wise supervised learning, however, there could exist a severe mismatch between the poor separability of feature space in shallow layers and the strong supervised constraint of local loss functions, as described below, 
\begin{equation}
\begin{aligned}
L_{i} = ||f_{i}(\mathbf{h}_{i})-Y||_{F}^{2}, i \in \{1,2,\dots, n\}
\end{aligned}
\end{equation}
where $i$ indicates the $i$\_th layer. From which, we can observe that both the global loss $L_{G}$ and the local loss $L_{i}$ adopt the same classification label set $Y$, which would bring in the mismatch we mentioned above and result in  the deviation of parameter optimization, specifically in shallow layers. 

The feature space becomes sparse, abstract, semantic with the network getting deeper. Thus, the separability of feature space is enhanced when the network goes deeper. From this perspective, the layer-wise learning networks should put intensive convolution operations in the latter part of the network so as to enhance feature extraction and learning ability, and relax the convolution intensity in the shallow part to get it matched with the feature separability.

\subsection{Downsampling Acceleration}
We propose an accelerated downsampling scheme to cope with the poor separability of feature space and improve the performance of layer-wise learning for deeper neural networks. Pooling operates on feature space for data compression by down-sampling feature maps in the way of summarising the presence in patches \cite{b35}. Thus, it is a way that is spatially related to feature selection. By consecutive pooling operations, the network realizes feature down-scaling and semantic extraction. While if the stride of pooling is radical, it may lead to feature missing and impair the entire ability of feature learning. Thus, according to the layer hierarchy of current deep neural networks, downsampling acceleration in space is not a feasible approach. 

To make the network focus more on features that are separable and to relax the mismatch of feature separability and strong supervision constraint in shallow layers, we propose to accelerate downsampling by more intensive pooling operations with the same but small stride in the bottom part of the network so as to reduce the convolution intensity in shallow layers. 
In this way, the deviation of learning optimization in shallow layers could be reduced and the contribution of the local regression in shallow layers could be more positive towards the ultimate classification performance. This is a method that would take full advantage of the natural characteristic of feature expression in the situation of the concatenated layer distribution. We name it the downsampling acceleration in the time domain. Given the layer-wise convolution operation set $F=[f_{1}, f_{2}, \dots, f_{n}]$, its corresponding set of convolution layer distribution $D=[d_{1}, d_{2}, \dots, d_{n}]$. If $k$ is the layer depth that defines the shallow layers, then we have,
\begin{equation}
\begin{aligned}
k(d_{i}) < k(d_{j}), k < \frac{n}{2}, i\leq k, k<j \leq n 
\end{aligned}
\end{equation}
\begin{equation}
\begin{aligned}
\arg\min_{D}Scale(X_{i}), i \leq k
\end{aligned}
\end{equation}
where $n$ is the layer number of the entire network, and function $Scale$ calculates the resolution scale of the feature map $X$ in the layer depth $i$. 

\section{Experimental Analysis}
We performed experiments on the typical image classification benchmarks, CIFAR10 \cite{b36} and CIFAR100 \cite{b36}, to evaluate the performance of the proposed method. We verify the ResNet architectures that researchers found hard to fit the layer-wise learning \cite{b6}. For each model, we compared the performance when trained with the global learning (i.e., standard global back propagation) and with the state-of-the-art local layer-wise learning (i.e., predsim loss \cite{b6}), where typically the convolution operation in the by-pass classification branch ($W_{\theta^{'}_{j}}$ in equation \eqref{eq2}) is one 3$\times$3 convolution. The execution details of the experiments are discussed below: 

\begin{itemize}
\item The hyper-parameters are kept identical for both learning scenarios for a given dataset and a specific architecture. Experiments are kept fairly competitive by merely varying the dropout, learning rate, and hidden layer dimension across different networks. 
\item The models are implemented using the Pytorch framework. The output layer is trained normally by the cross-entropy loss function for both learning scenarios.
\item A batch size of 128 is applied to all the experiments. The learning solver is ADAM \cite{b37} for optimization. 
\item For both learning scenarios, the networks are trained using leaky-ReLU non-linearity with the negative slope of 0.01 \cite{b6} for the consideration of training stability. We use the same sequence of batch normalization, non-linearity, dropout with an equal rate for all layers.
\item There are 400 training epochs for both CIFAR10 and CIFAR100 experiments. The learning rate is multiplied by the decay factors of 0.25 at 50\%, 75\%, 89\%, and 94\% of the total training epochs.
\end{itemize}

\subsection{CIFAR10}
The CIFAR10 dataset consists of 50,000 training images with a resolution of 32$\times$32  pixels \cite{b36}. It has 10 image classes. The initial learning rate for all the ResNet networks in Table \ref{table:experiment_cifar10} and Table \ref{table:experiment_cifar100} is 3e-4. The dropout rate is 0.3 accordingly. The ``Layer-wise'' term here refers to the network trained by layer-wise learning method with the state-of-the-art Predsim loss \cite{b6}. 

As a baseline, we include the classification performance of the original ResNet18 and ResNet50 in the tables. Note that the best layer-wise result of ResNet18 family is reported with the hierarchical structure of [1,1,2,2], which has less shallow layers and more deep layers. The comparison between ResNet\_r2 and ResNet\_r3 shows that more intensive convolution operation in shallow layers leads to worse performance while the comparison between ResNet18\_r1 and ResNet18\_r4, indicating the deeper convolutions in the deeper part of the network brings about equal or even better results. From this perspective, downsampling acceleration (i.e., ResNet18\_r4) weakens the poor parameter learning in shallow layers and increases the expression ability of deeper layers. This can also be verified in the case of ResNet50, where ResNet50\_r4 and ResNet50 hold the same number of convolution layers while ResNet50\_r4, after downsampling acceleration, increases the classification performance by a large margin of over 4\% and it is the method that obtains the least margin of performance with the corresponding global learning method by roughly 0.5\%.

\begin{figure}
    \includegraphics[width=0.95\linewidth]{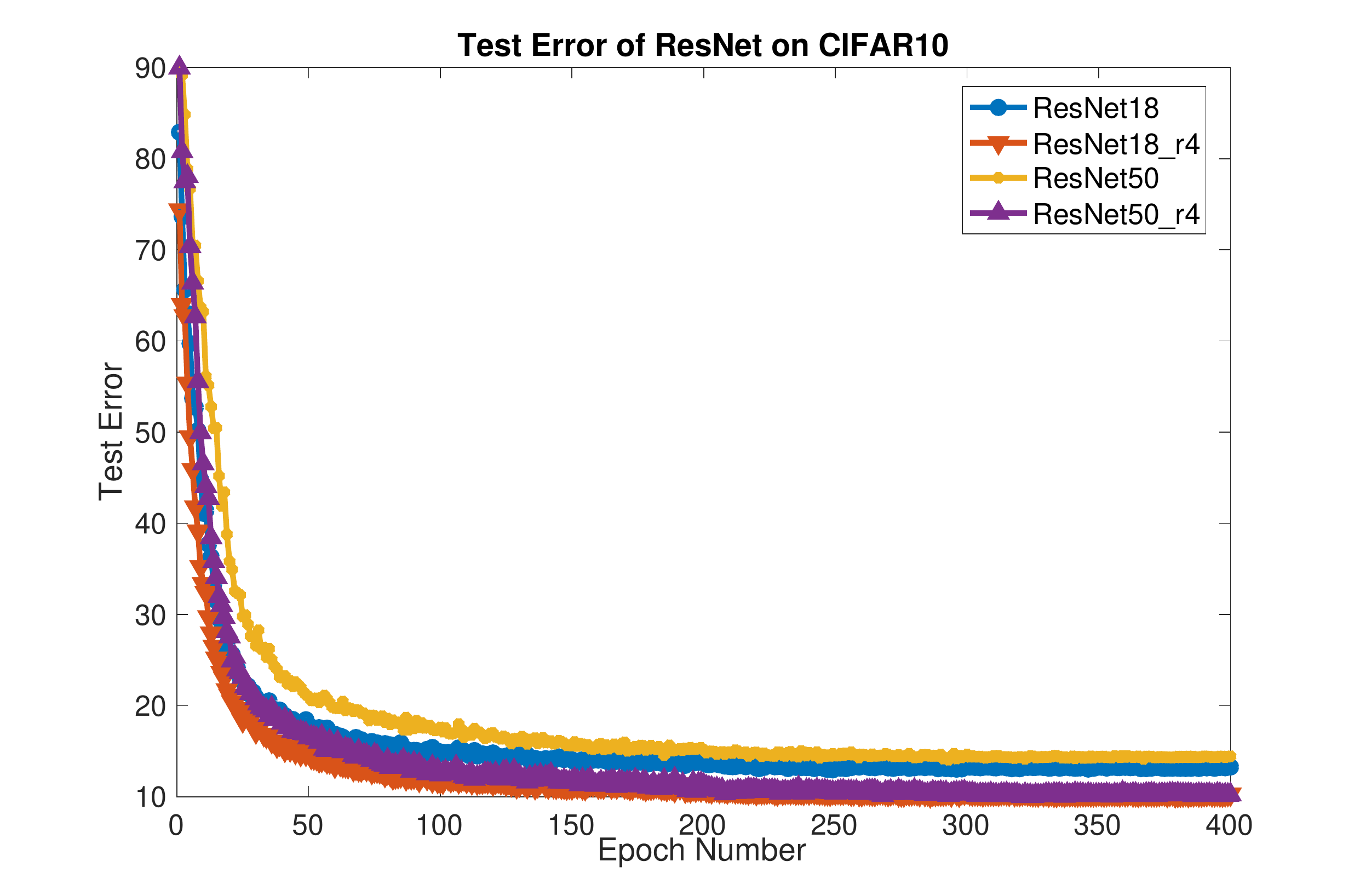}\hfill
    \vspace{-2mm}
    \caption{Layer-wise Test Error of ResNet on CIFAR10.}
	\label{fig:cifar10}
 	\vspace{-3mm}
\end{figure}

From the test error curves during the inference time in Figure \ref{fig:cifar10}, we can see that the performance of both ResNet18 and ResNet50 trained in the way of layer-wise learning fall behind that of ResNet18\_r4 and ResNet50\_r4 throughout the entire inference procedure (the purple and red curves are always under the yellow and blue ones), where the latter two models further employ the downsampling acceleration. We can observe that the test errors of models with pooling positioned ahead have a sharper drop in the first 100 epochs than the other two, which may be due to the faster convergence brought by the usage of downsampling acceleration. Finally, the red and purple curves rest on a state that is clearly lower than the yellow and blue ones, showing the clear advantage of downsampling accelerated ResNet models over the ones with simple layer-wise learning. 
\setlength{\tabcolsep}{4pt}
\begin{table}
\begin{center}
\caption{Experiment results with ResNets on CIFAR10.}
\label{table:experiment_cifar10}
\begin{tabular}{|c|c|c|c|c|c|}
\hline
Model & LR & DR & Hierarchy & Layer-wise & Global\\
\hline
ResNet18 & 3e-4 & 0.3 &[2,2,2,2] & 86.94 & 92.45\\
\hline
ResN18\_r1 & 3e-4 & 0.3 & [1,1,1,1] & \underline{89.26} & 92.75\\
\hline
ResNet18\_r2 & 3e-4 & 0.3 & [2,2,1,1] & 86.27 & 92.39\\
\hline
ResNet18\_r3 & 3e-4 & 0.3 & [2,1,1,1] & 87.61 & 92.48\\
\hline
ResNet18\_r4 & 3e-4 & 0.3 & [1,1,2,2] & \textbf{89.77} & 93.01\\
\hline
ResNet50 & 3e-4 & 0.3 & [3,4,6,3] & 85.6 & 89.65\\
\hline
ResNet50\_r1 & 3e-4 & 0.3 & [2,3,6,3] & 87.62 & 90.73\\
\hline
ResNet50\_r2 & 3e-4 & 0.3 & [1,2,6,3] & \underline{89.68} & 91.69\\
\hline
ResNet50\_r3 & 3e-4 & 0.3 & [2,3,7,4] & 87.93 & 90.52\\
\hline
ResNet50\_r4 & 3e-4 & 0.3 & [1,2,8,5] & \textbf{89.81} & 90.49\\
\hline
\end{tabular}
\end{center}
\vspace{-2mm}
\end{table}
\setlength{\tabcolsep}{1.4pt}

\setlength{\tabcolsep}{4pt}
\begin{table}
\begin{center}
\caption{Experiment results with ResNets on CIFAR100.}
\label{table:experiment_cifar100}
\begin{tabular}{|c|c|c|c|c|c|}
\hline
Model & LR & DR & Hierarchy & Layer-wise & Global\\
\hline
ResNet18 & 3e-4 & 0.1 &[2,2,2,2] & 66.05 & 73.13\\
\hline
ResN18\_r1 & 3e-4 & 0.1 & [1,1,1,1] & \underline{66.37} & 73.6\\
\hline
ResNet18\_r2 & 3e-4 & 0.1 & [2,2,1,1] & 64.54 & 74.08\\
\hline
ResNet18\_r3 & 3e-4 & 0.1 & [2,1,1,1] & 64.84 & 73.55\\
\hline
ResNet18\_r4 & 3e-4 & 0.1 & [1,1,2,2] & \textbf{68.32} & 73.76\\
\hline
ResNet18\_r5 & 3e-4 & 0.1 & [1,1,2,4] & 69.12 & 72.8\\
\hline
ResNet50 & 3e-4 & 0.1 & [3,4,6,3] & 67.01 & 72.34\\
\hline
ResNet50\_r1 & 3e-4 & 0.1 & [2,3,6,3] & 68.64 & 73.94\\
\hline
ResNet50\_r2 & 3e-4 & 0.1 & [1,2,6,3] & \textbf{69.53} & 73.86\\
\hline
ResNet50\_r3 & 3e-4 & 0.1 & [2,3,7,4] & 68.06 & 72.82\\
\hline
ResNet50\_r4 & 3e-4 & 0.1 & [1,2,8,5] & \underline{68.98} & 72.77\\
\hline
\end{tabular}
\end{center}
\vspace{-2mm}
\end{table}
\setlength{\tabcolsep}{1.4pt}

\begin{figure}
    \includegraphics[width=0.95\linewidth]{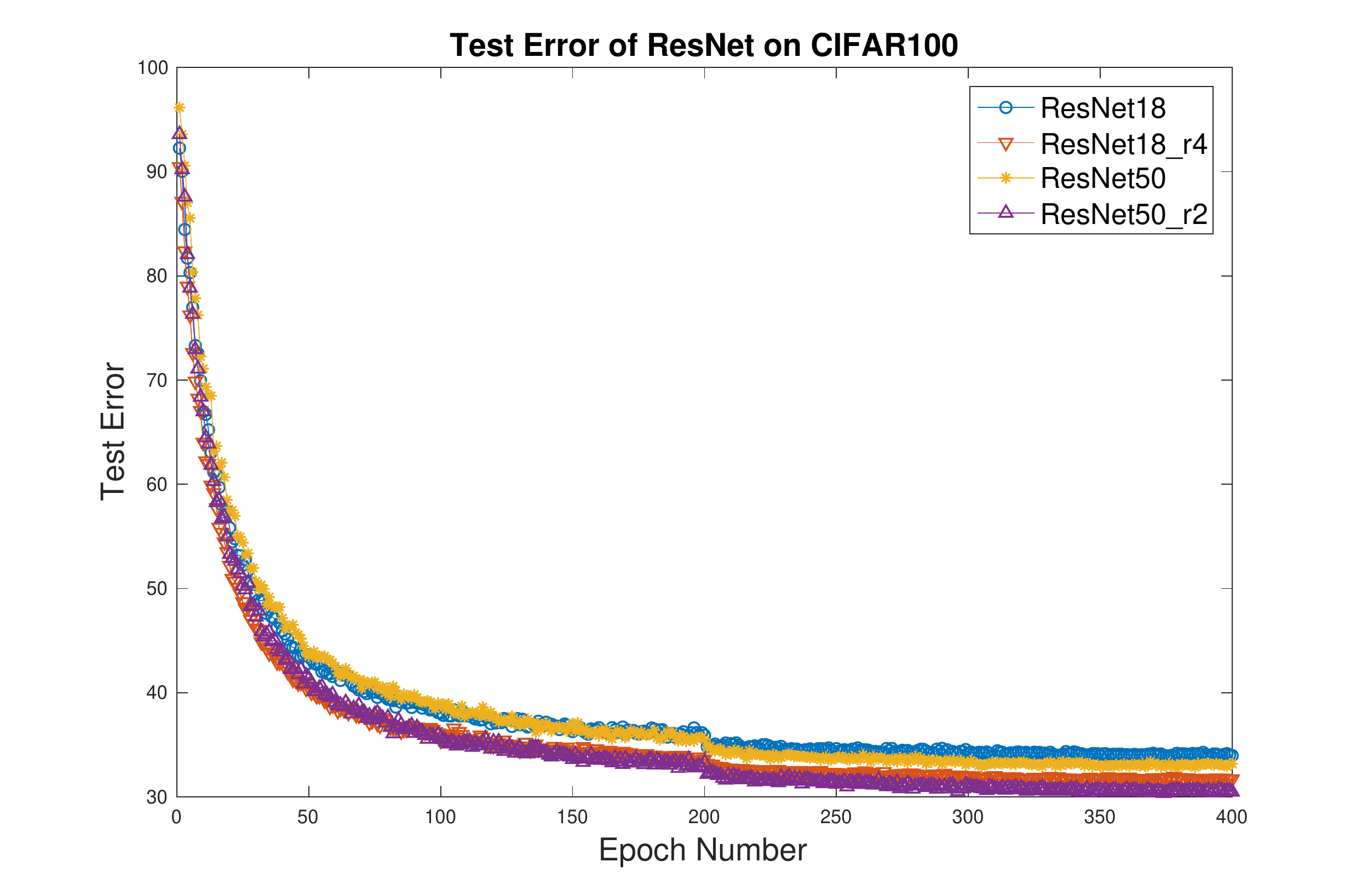}\hfill
    \vspace{-2mm}
    \caption{Layer-wise Test Error of ResNet on CIFAR100.}
	\label{fig:cifar100}
 	\vspace{-3mm}
\end{figure}

\subsection{CIFAR100}
CIFAR100 consists of 50,000 training images with a resolution of 32$\times$32 pixels \cite{b36}. It has 100 classes. The initial learning rate in Table \ref{table:experiment_cifar100} is 3e-4. The dropout rate is 0.1. 

As the baseline, we implemented and tested the classification accuracy of both original ResNet18 and ResNet50. Note that the best results appear in the cases of ResNet18\_r5 and ResNet50\_r2, both are making the full advantage of downsampling acceleration. The comparison between ResNet18\_r1 and ResNet18\_r3 verifies that the downsampling acceleration has a positive impact on improving the poor separability of shallow layers. ResNet18 and ResNet18\_r5 have the same number of convolution layers while ResNet18\_r5 leads the former by nearly 7\%, which strongly demonstrates that the downsampling acceleration could effectively avoid the influence by poor separability of shallow features and push the network to focus on deeper features with better semantic expression and separability so as to realize the matching between feature separability and strong supervised constraint.

Accordingly, ResNet50\_r2 and ResNet50\_r4 obtain better results compared with the base model of ResNet50 by a margin of 2\% by introducing the downsampling acceleration. We draw the test error curve of ResNet50\_r2 in Figure \ref{fig:cifar100} with regards to its best accuracy in this situation for better comparison. From Figure \ref{fig:cifar100}, we can see that the performances of models with pooling positioned ahead, ResNet18\_r4 and ResNet50\_r2 maintain the same advantage over the other two with simple layer-wise learning as that on CIFAR10 in Figure \ref{fig:cifar10}. This also demonstrates the generalization ability of the proposed downsampling acceleration scheme. From the above analysis, we can conclude that the downsampling acceleration can be widely applicable to different network models. 

\section{Conclusion}
In this paper, we have explored the fundamental reason that impedes the development of current state-of-the-art layer-wise models towards deeper scales is due to the poor separability of feature space in shallow layers. To overcome this limitation, we have proposed a customized approach by downsampling acceleration so as to equip the layer-wise learning with a better matching of separability and supervision restriction. Extensive experiments have demonstrated the advantages of employing the proposed downsampling acceleration and the performance is improved with the state-of-the-art layer-wise approaches on deeper ResNet family. The study makes it possible to apply the layer-wise learning to deeper networks with restricted resources or efficiency requirements. Generally, the study reveals the mechanism that underlies the development of layer-wise learning towards complicated cases and provides a simplification approach for future theoretical research in the layer-wise based learning.




\balance

\balance

\end{document}